\documentclass[9pt,a4paper]{article}

\usepackage[T1]{fontenc}
\usepackage{tgheros}

\usepackage{newunicodechar}
\newunicodechar{–}{--}
\newunicodechar{—}{---}
\newunicodechar{’}{'}
\newunicodechar{“}{``}
\newunicodechar{”}{''}

\usepackage[margin=2.5cm]{geometry}
\usepackage{titling}
\pretitle{\centering}
\posttitle{\par\vskip 9.5pt}
\preauthor{\par\centering\fontsize{10pt}{14.4pt}\selectfont}
\postauthor{\par}
\predate{}\postdate{}

\usepackage{setspace}
\usepackage{graphicx}
\usepackage{float}
\usepackage{lipsum}
\usepackage[dvipsnames]{xcolor}
\definecolor{gfaiblau}{HTML}{23355D}
\usepackage{sectsty}
\usepackage{hyperref}
\usepackage{booktabs,multicol,multirow,tabularx,array}    
\usepackage[withpage]{acronym}
\usepackage{cite}
\usepackage{shortvrb}
\usepackage{pdflscape}

\usepackage{comment}

\fontsize{9pt}{12pt}\selectfont

\subsectionfont{\centering\color{gfaiblau}\fontsize{10pt}{12pt}\selectfont\bfseries}
\sectionfont{\centering\color{gfaiblau}\fontsize{10pt}{12pt}\selectfont\bfseries}
\makeatletter
\renewcommand{\@seccntformat}[1]{\csname the#1\endcsname\hspace{0.3em}}
\renewcommand\normalsize{\@setfontsize\normalsize{9pt}{14.4pt}}
\renewcommand\section{\@startsection{section}{1}{\z@}%
      {-3.5ex \@plus -1ex \@minus -.2ex}{3.4ex \@plus.2ex}%
      {\normalfont\Large\bfseries\SS@sectfont}}

\makeatother
\normalsize
\newlength{\preSecSkip}
\let\oldsection\section
\renewcommand{\section}{\vspace*{\preSecSkip}\oldsection}

\title{\color{Black}{\bfseries\fontsize{12pt}{14pt}\selectfont Photogrammetric 3D Model-Based Animal Behavior Classification}}
\author{%
{Hellwich, Olaf}$^{1,5}$, {Heine, Moritz}$^{1}$, {Reiske, Patrik}$^{1,5}, ${Traverso, Sole}$^{3,1,5}$, {Hohlbaum, Katharina}$^{2,5}$, \\ {Sprekeler, Henning}$^{3,5}$, {Thöne-Reineke, Christa}$^{4,5}$, {Lewejohann, Lars}$^{2,4,5}$, {Bhuiyan, Roman}$^{1}$ \\
$^1$TU Berlin, Computer Vision \& Remote Sensing \\
$^2$German Centre for the
Protection of Laboratory Animals (Bf3R), \\ German Federal Institute for Risk Assessment (BfR) \\
$^3$TU Berlin, Modeling of Cognitive Processes \\
$^4$FU Berlin, Institute of Animal Welfare, Animal Behavior and Laboratory Animal Science \\
$^5$TU Berlin, Science of Intelligence Excellence Cluster
}
\begin{document}
\date{}
\maketitle
\thispagestyle{empty}
\vspace{9pt}
\noindent
{\fontsize{9pt}{14.4pt}\selectfont
{\bfseries\color{gfaiblau}\fontsize{10pt}{14.4pt}\selectfont Abstract:}
Video-based behavior estimation is often implemented end-to-end from image sequences. We advocate a two-stage alternative: reconstruct 3D body and object geometry before classifying behavior. This makes ambiguities of the ill-posed 2D-to-3D inverse problem explicit and potentially correctable. We demonstrate the strategy with simulated interacting mice and extend it to real video.
}

\vspace{15pt}
{\centering
\fontsize{9pt}{14.4pt}\selectfont
{\bfseries\color{gfaiblau}\fontsize{10pt}{14.4pt}\selectfont Keywords:}
Video-based animal observation; 3D motion reconstruction; Behavior classification; Behavior simulation.}

\section{Introduction}

Behavior is a higher-level construct inferred from measurable movement patterns. In a semantic network, video features may connect directly to movement, whereas behavior is inferred indirectly through movement \cite{hellwich2000object}.
Because video provides only 2D projections and cannot capture all causal subtleties of living behavior, behavior is not directly observable; inferring it from video is therefore an ill-posed inverse problem.
We therefore propose a fully three-dimensional (3D) simulation and reconstruction framework: 3D body kinematics are reconstructed from video first, and behavioral states are inferred afterward rather than directly from raw frames.

Reconstruction also uses prior 3D models: deformable animal bodies, static cage geometry, and kinematic enrichment devices such as lock-boxes. These provide structural and motion constraints.
We illustrate the concept with simulated interactions between two mice and real multi-view video of a mouse solving a lock-box task.

\section{State-of-the-Art of Animal Behavior Classification}

Modern video-based animal behavior pipelines divide activity into discrete, functionally meaningful classes. They typically track body parts with pose estimators such as DeepLabCut \cite{mathis2018deeplabcut}, then model behavior from the resulting trajectories using supervised or unsupervised temporal analysis. Traditional systems mainly use monocular 2D keypoints, while newer methods increasingly accept reconstructed 3D trajectories.

MARS recognized mouse social behaviors from video~\cite{mars2021}; B-SOiD discovered behavioral motifs unsupervised from pose trajectories~\cite{bsoid2021}; and A-SOiD added active learning to reduce annotation while improving classification~\cite{asoid2024}. Although commonly demonstrated on 2D data, B-SOiD can in principle incorporate 3D trajectories.

VAME learns latent motion representations with variational sequence models and supports 2D or reconstructed 3D trajectories~\cite{vame2022}. SimBA provides supervised, feature-based behavior classification from pose outputs and can likewise accommodate 3D keypoints~\cite{simba2024}.

Keypoint-MoSeq extends MoSeq to pose-estimation keypoints and naturally supports 2D and 3D skeletons~\cite{keypointmoseq2024}. DLC2Action performs temporal action recognition from DeepLabCut-style trajectories and is similarly compatible with both 2D and 3D settings~\cite{dlc2action2025}.

\section{Example A: 3D Reconstruction and Behavior Classification of Two Interacting Simulated Mice}

Two simulated mice exhibit running, rearing, and intermittent social interaction in an enclosed environment. \footnote{The code for this example was created from early February 2026 until mid-April 2026 using ChatGPT Plus 5.2 to 5.5. It is made available at git@github.com:hellwich/mouseModelPaper.git including the LLM chat history.}

\subsection{Mechanistic Simulation of Two Interacting Mice}

Each simulated mouse is an autonomous agent with a planar trajectory, discrete behavior (running, rearing, or social interaction), a keypoint skeleton, and non-overlap collision handling. Its 3D keypoints are projected into virtual camera images.
Control is hierarchical: hard physical and geometric constraints take precedence, followed by social objectives and then individual running or rearing. Social interaction therefore overrides individual behavior when necessary.

The body is a 3D keypoint graph whose nodes are anatomical landmarks and whose edges encode skeletal connectivity. Each trajectory stores planar position, heading, speed, and yaw rate; speed and yaw are damped and noise-perturbed, then adjusted for wall repulsion, social attraction, and inter-mouse avoidance. The mouse then advances along its heading, producing smooth stochastic motion with bounded speed and turning.

Individual states are running or rearing, with timers, gait phase, and rearing parameters. Running lasts 2--6 s and rearing 0.6--1.6 s; after running, rearing begins with probability 0.25, while rearing always returns to running. During social episodes, these timers are frozen.
Running adds sinusoidal head jitter and a trot-like triangular paw cycle with diagonal pairs in phase; gait phase scales with speed and stride length. Rearing rotates the skeleton around the tail root, fixes the hind paws, and repositions the forepaws. This procedural pose ignores dynamics such as balance, torques, and joint limits.

Collision handling approximates each body by front and rear 2D spheres derived from the keypoint graph. After movement, projection keeps spheres inside the cage and separates overlapping mice, with dominance weights controlling how much each mouse is displaced.

Social interaction is a pair-level finite-state machine with ``none'', ``approach'', ``hold'', and ``separate'' phases. Its state records the phase, initiator, meeting point, touch time, and minimum nose distance. Episodes are scheduled stochastically with a cool-down; the initiator animal approaches more directly than the responder, close noses trigger holding, and a fixed hold is followed by separation until sufficient distance is reached.

Per frame, arbitration updates the social state, individual timers when non-social, steering, trajectories, hard constraints, local and world skeleton poses, and finally 3D keypoints and ethogram labels. The skeleton is projected into virtual cameras to produce the synthetic video (Fig. \ref{fig:topfrontside}, top left).

\subsection{Video Rendering}

\begin{figure*}[htbp]
  \centering
  \begin{minipage}[t]{0.35\textwidth}
    \centering
    \includegraphics[width=\textwidth]{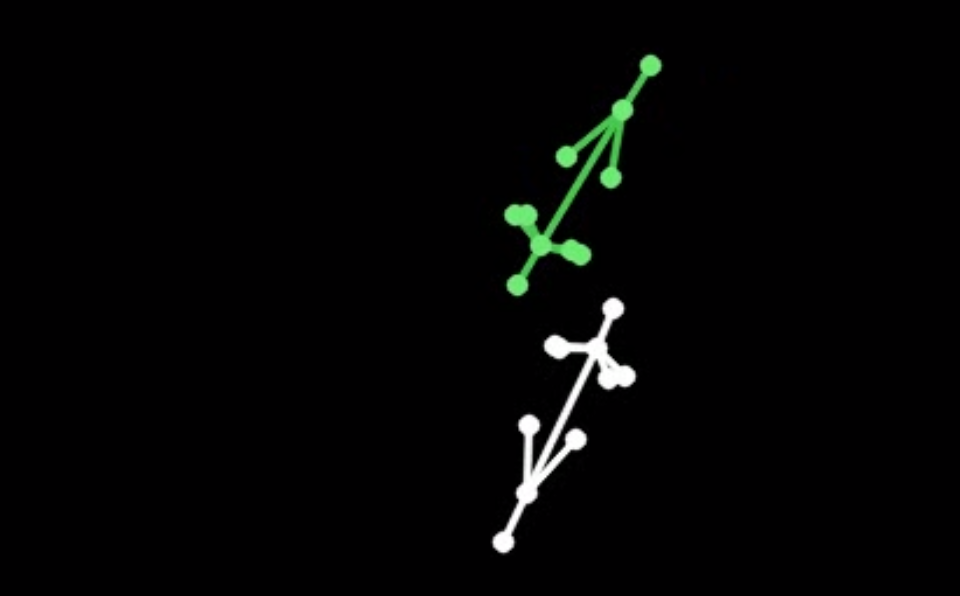}\\[0.5em]
    \includegraphics[width=\textwidth]{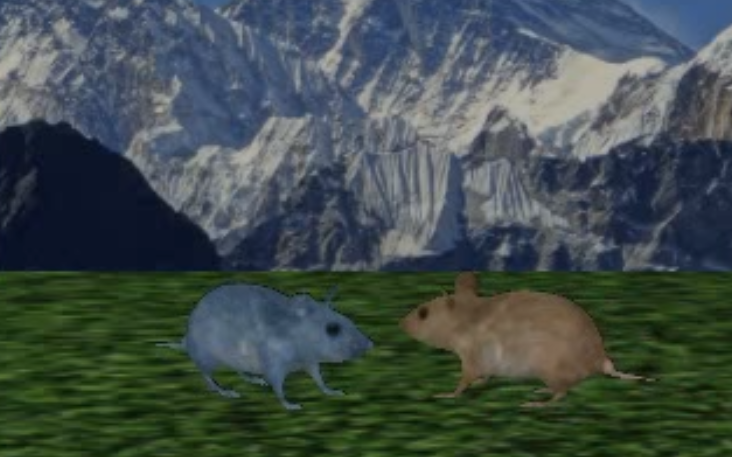}
  \end{minipage}\hspace{0.05\textwidth}
  \begin{minipage}[t]{0.35\textwidth}
    \centering
    \includegraphics[width=\textwidth]{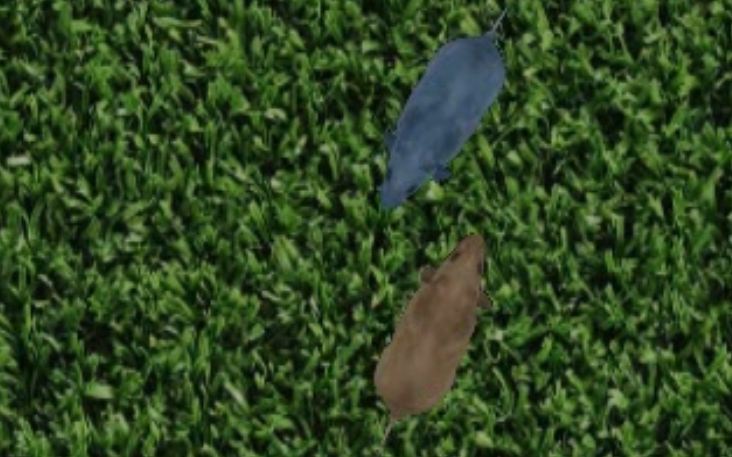}\\[0.5em]
    \includegraphics[width=\textwidth]{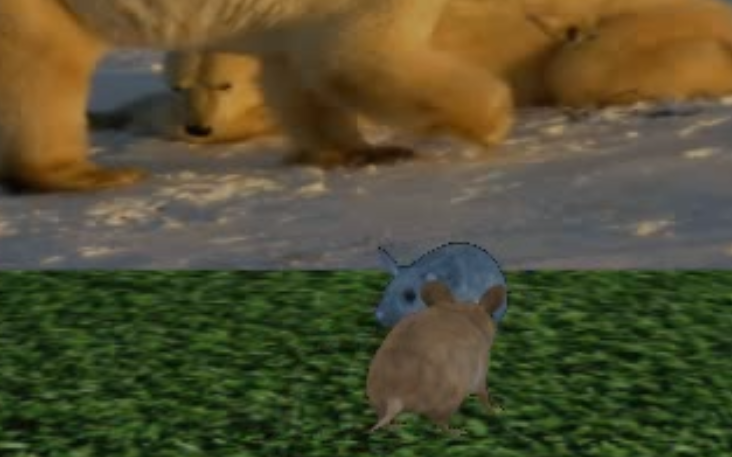}
  \end{minipage}
  \vspace{6pt}
  \caption{Simulated skeleton and rendered video. Top, front and side view}
  \label{fig:topfrontside}
\end{figure*}

Cage surfaces are rendered by homographically transforming textured images. Mouse bodies use a textured mesh deformed by a 3D rubber-sheet transformation from corresponding skeletal and mesh landmarks. Videos are rendered from three orthogonal views (Fig. \ref{fig:topfrontside}, top right, and bottom row).

\subsection{Extraction of 2D Body-Part Keypoints and 2D Body Segments}

\begin{figure*}[!tb]
  \centering
  \begin{minipage}[t]{0.32\textwidth}
    \centering
    \includegraphics[width=\textwidth]{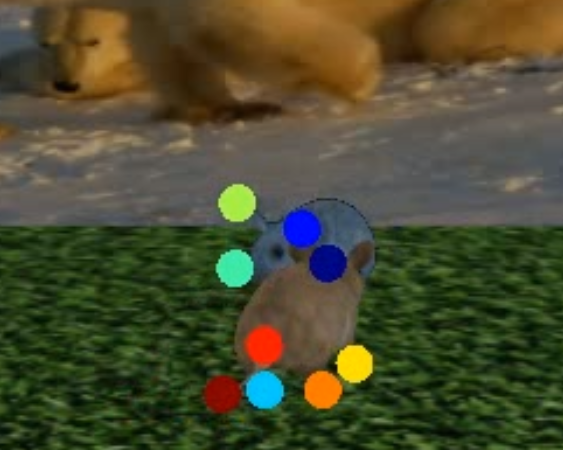}\\[0.5em]
    \includegraphics[width=\textwidth]{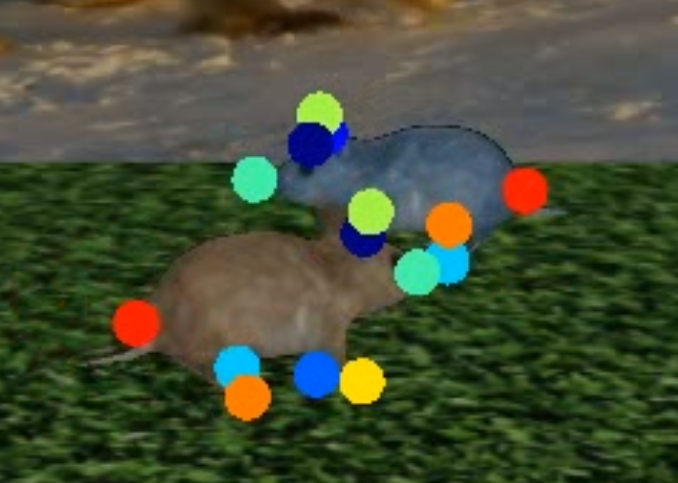}
    \caption{DLC keypoints}
    \label{fig:DLCkeypoints}
  \end{minipage}\hfill
  \begin{minipage}[t]{0.32\textwidth}
    \centering
    \includegraphics[width=\textwidth]{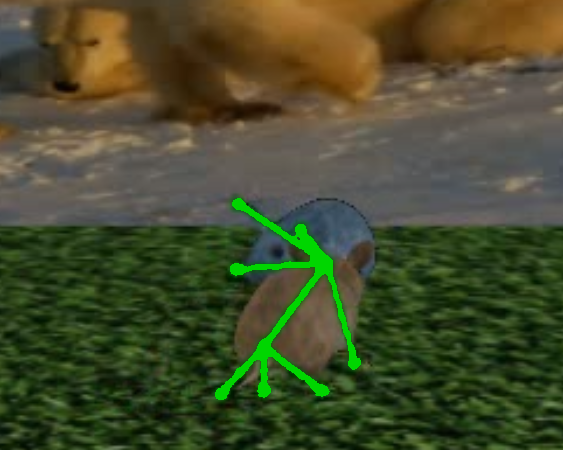}\\[0.5em]
    \includegraphics[width=\textwidth]{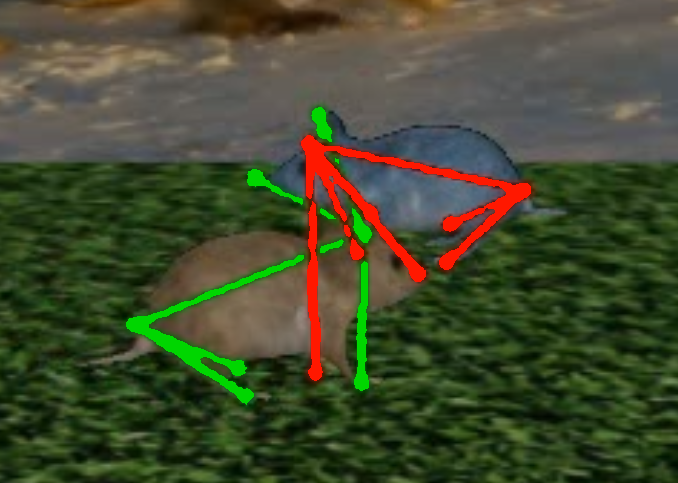}
    \caption{The same DLC keypoints with skeleton graphs}
    \label{fig:DLCskeletons}
  \end{minipage}\hfill
  \begin{minipage}[t]{0.32\textwidth}
    \centering
    \includegraphics[width=\textwidth]{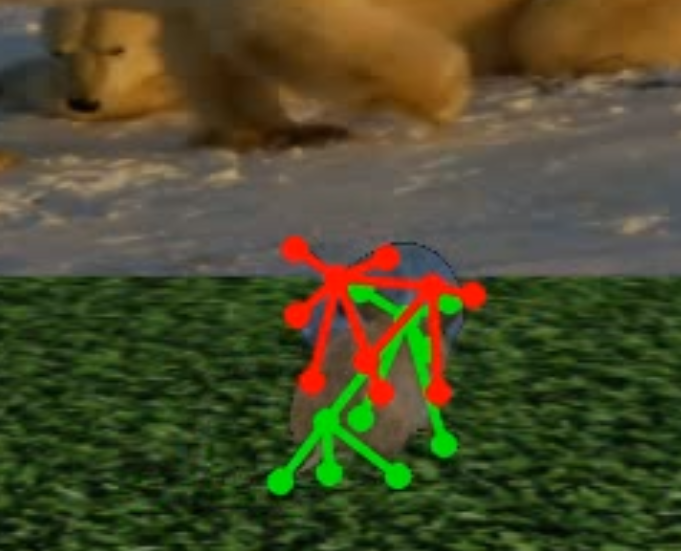}\\[0.5em]
    \includegraphics[width=\textwidth]{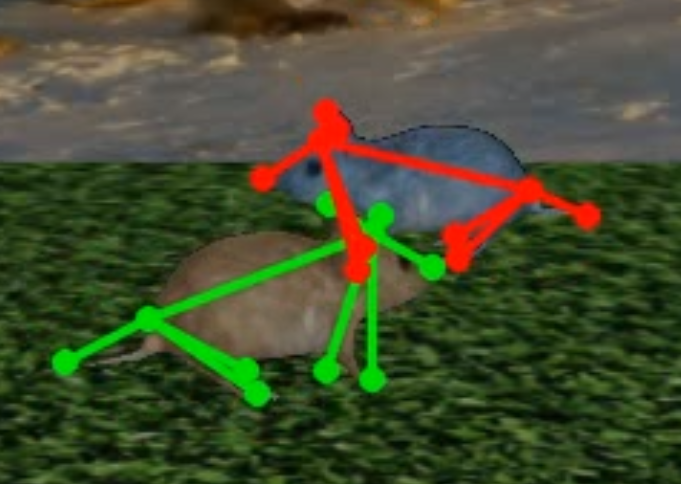}
    \caption{3D  skeletons after keypoint cleaning}
    \label{fig:cleanedDLCskeletons}
  \end{minipage}
\end{figure*}

DeepLabCut (multi-animal maDLC, version 2.3.11, \cite{lauer2022multi}) was trained on arbitrarily generated simulation images from top, front, and side views. This synthetic setting permits essentially unlimited labeled training data. Under mutual occlusion (Fig. \ref{fig:DLCkeypoints}), extracted skeletons can connect parts from different mice (Fig. \ref{fig:DLCskeletons}), while identities may flip over time and disagree across views.

\begin{figure*}[!tb]
    \centering
    \begin{minipage}[t]{0.35\textwidth}
        \centering
        \includegraphics[width=\textwidth]{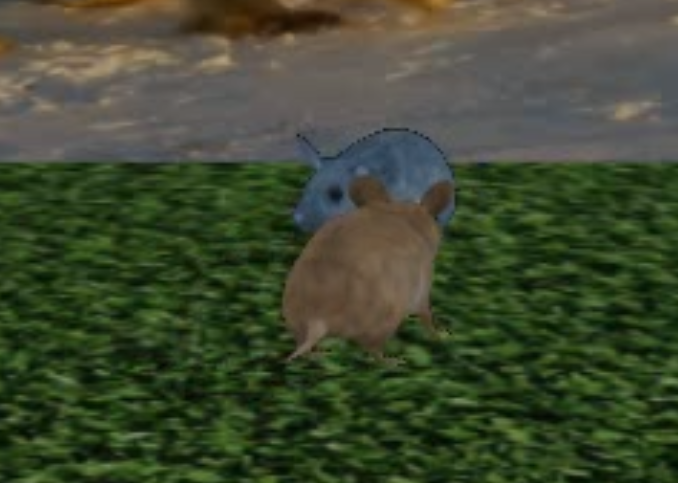}
        \caption{Overlapping 2D body segments}
        \label{fig:FrameUnetSeg}
    \end{minipage}\hspace{0.05\textwidth}%
    \begin{minipage}[t]{0.35\textwidth}
        \centering
        \includegraphics[width=\textwidth]{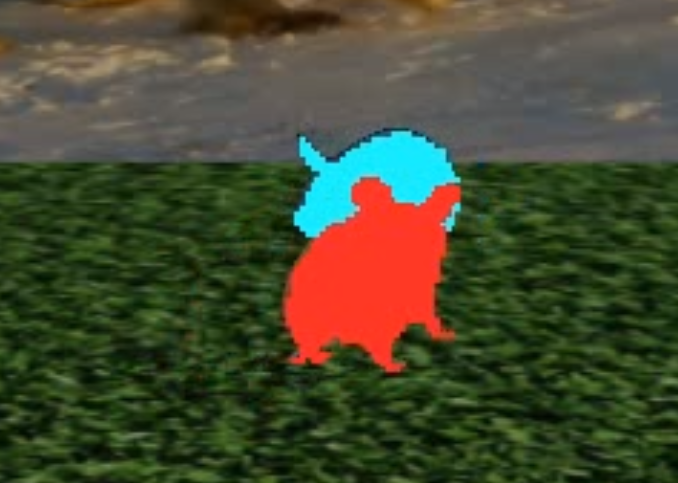}
        \caption{U-net semantic segmentation of overlapping bodies}
        \label{fig:UnetSeg}
    \end{minipage}
\end{figure*}

Mouse body segmentation was also learned from rendered masks for background, mouse 0, and mouse 1. A multi-class 2D encoder-decoder U-Net with GroupNorm and previous-mask conditioning was trained \cite{ronneberger2015unet}. Example predictions for the image in Fig. \ref{fig:FrameUnetSeg} are shown in Fig. \ref{fig:UnetSeg}.

\subsection{3D Reconstruction From Multi-View 2D Observations, 3D-Model and Temporal Constraints}
\label{sec:datacleaning}

Multi-stage reconstruction combines calibrated cameras, per-frame 2D keypoints, and segmentation masks to produce cleaned 3D trajectories and corrected identities for both mice.

For every body part and frame, all plausible two- and three-camera detection combinations are tested, regardless of their original mouse labels. Identity inconsistencies across views are therefore resolved by selecting combinations that are geometrically plausible in 3D.

Stage A validates 3D hypotheses from calibrated viewing rays. Candidate pairs or triplets must have sufficiently non-degenerate intersection angles, pass near one another in 3D, and reproject accurately into the source images. Accepted points are refined by Ceres single-point bundle adjustment \cite{Agarwal_Ceres_Solver_2022}, with plausible three-view triplets preferred over pairs. A linearized least-squares extension can additionally impose a deformable-mouse motion-track constraint \cite{hellwich2024tracking}.
Per body part, up to two disjoint 3D hypotheses are then selected, so no 2D detection is allowed to contribute to both mice. Untriangulated detections remain as single-view observations for later assignment.

Stage B assigns accepted 3D points and remaining single-view detections to mouse 0 or mouse 1 using multiple identity cues.
First, semantic segmentation provides local identity votes for each 2D keypoint. Presence, rather than pixel area size, determines support; so a small visible ear can identify the foreground mouse even when a larger occluded body lies behind it. Votes are accumulated across contributing views, and exclusive support yields an identity cue. Second, temporal continuity repairs frame-to-frame identity flips.

Third, assigned head and tail-root anchors define a centroid and body axis for each mouse. Other 3D parts are assigned by a cost combining segmentation exclusivity, temporal continuity, distance to the centroid, and distance to the body axis, allowing geometric evidence to override the original DLC identity.

Fourth, the prior 3D skeleton checks template-relative edge lengths and alternative head/tail-root pairings to prevent chimeric ``half-stitch'' bodies combining parts of different mice. Violating skeleton leaf parts may be freed, reassigned, or dropped, while rigid template fitting provides a fallback plausibility test when enough parts are available. Finally, remaining single-view detections are assigned.

This cleaning yields 3D skeleton tracks comparable to the simulated ground truth, although still incomplete and imperfect. Fig. \ref{fig:cleanedDLCskeletons} shows results for the preceding DLC examples.

\subsection{Behavior Classification from 3D Body-Part Tracks}

Both 3D skeleton tracks are converted to normalized, pair-centered temporal features used by sequence models to predict interaction and per-animal rearing, producing a framewise ethogram.

Coordinates are expressed in a local pair-centered co-ordinate system defined by the tail roots, with x directed from mouse 0 to mouse 1 and z vertical, reducing sensitivity to global position and rotation. Velocities and nose-to-nose and tail-root distances are added.

The classification net uses three binary heads: interaction, rearing of mouse 0, and rearing of mouse 1.
LSTM, xLSTM, and Transformer models operate on sliding temporal windows rather than isolated frames. Each uses a sequence backbone followed by the same three binary behavior heads.

Table \ref{tab:f1_behavior_comparison} reports F1 scores in a three-minute test sequence. The classification of reconstructed 3D tracks closely matches the classification of simulated ground-truth tracks, suggesting that the reconstruction pipeline preserves much of the information needed to distinguish evaluated behaviors despite ambiguous and incomplete 2D observations.

An end-to-end Transformer trained on perfect simulated multiview 2D projections, using pair-centered coordinates, performs substantially worse when evaluated on DLC keypoints from the same test sequence. Poor rearing performance also persists when this model receives perfect simulated test projections, indicating that DLC detection and identity errors alone do not explain the gap. These results support the effectiveness of the explicit 3D-reconstruction-based pipeline in the investigated setting but do not establish that 3D reconstruction is necessary or that a better-designed end-to-end model could not achieve comparable performance.

\begin{table}[ht]
\centering
\begin{tabular}{llccc}
\hline
Model & Input source & Interaction F1 & Rear0 F1 & Rear1 F1 \\
\hline
LSTM & Simulated 3D tracks & 0.6276 & 0.8369 & 0.8316 \\
 & Reconstructed 3D tracks & 0.6147 & 0.8116 & 0.8066 \\
xLSTM & Simulated 3D tracks & 0.7146 & 0.7500 & \textbf{0.8866} \\
 & Reconstructed 3D tracks & 0.6702 & 0.6776 & 0.8293 \\
Transformer & Simulated 3D tracks & \textbf{0.7271} & \textbf{0.8611} & 0.8743 \\
 & Reconstructed 3D tracks & \textbf{0.7240} & \textbf{0.8286} & \textbf{0.8432} \\
\hline
 & End-to-end multiview DLC 2D keypoints & 0.4071 & 0.0960 & 0.1351 \\
\hline
\end{tabular}
\vspace{6pt}
\caption{F1 scores for behavior classification on the same three-minute simulated test sequence. The 3D classifiers were trained on simulated 3D body-part tracks and evaluated on simulated or cleaned, completed 3D-reconstructed tracks. The direct Transformer was trained on perfect multiview 2D projections of the same behavior-training trajectory, using the same temporal training/validation split, and evaluated on DLC 2D detections. Its inputs use pair-centered image coordinates, without segmentation, explicit 3D reconstruction, or additional identity correction. Rearing scores are evaluated on ground-truth non-interaction frames. Bold values denote the highest score for each behavior within each of the two 3D input-source groups.}
\label{tab:f1_behavior_comparison}
\end{table}

\section{Example B: Empirical Observation of a Live Mouse Performing a Lock-Box Problem-Solving Task}

The second example is not yet fully evaluated with the two-stage pipeline; it instead illustrates weaknesses of video behavior classification without intermediate 3D reconstruction.

\begin{figure*}[htbp]
  \centering
  \begin{minipage}[t]{0.35\textwidth}
    \centering
    \includegraphics[width=\textwidth]{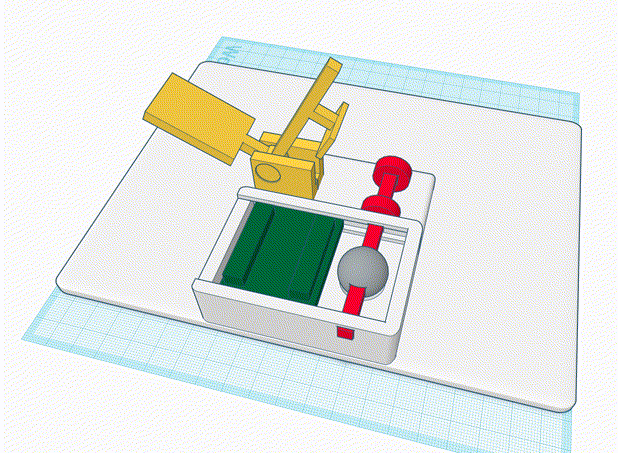}\\[0.5em]
    \includegraphics[width=\textwidth]{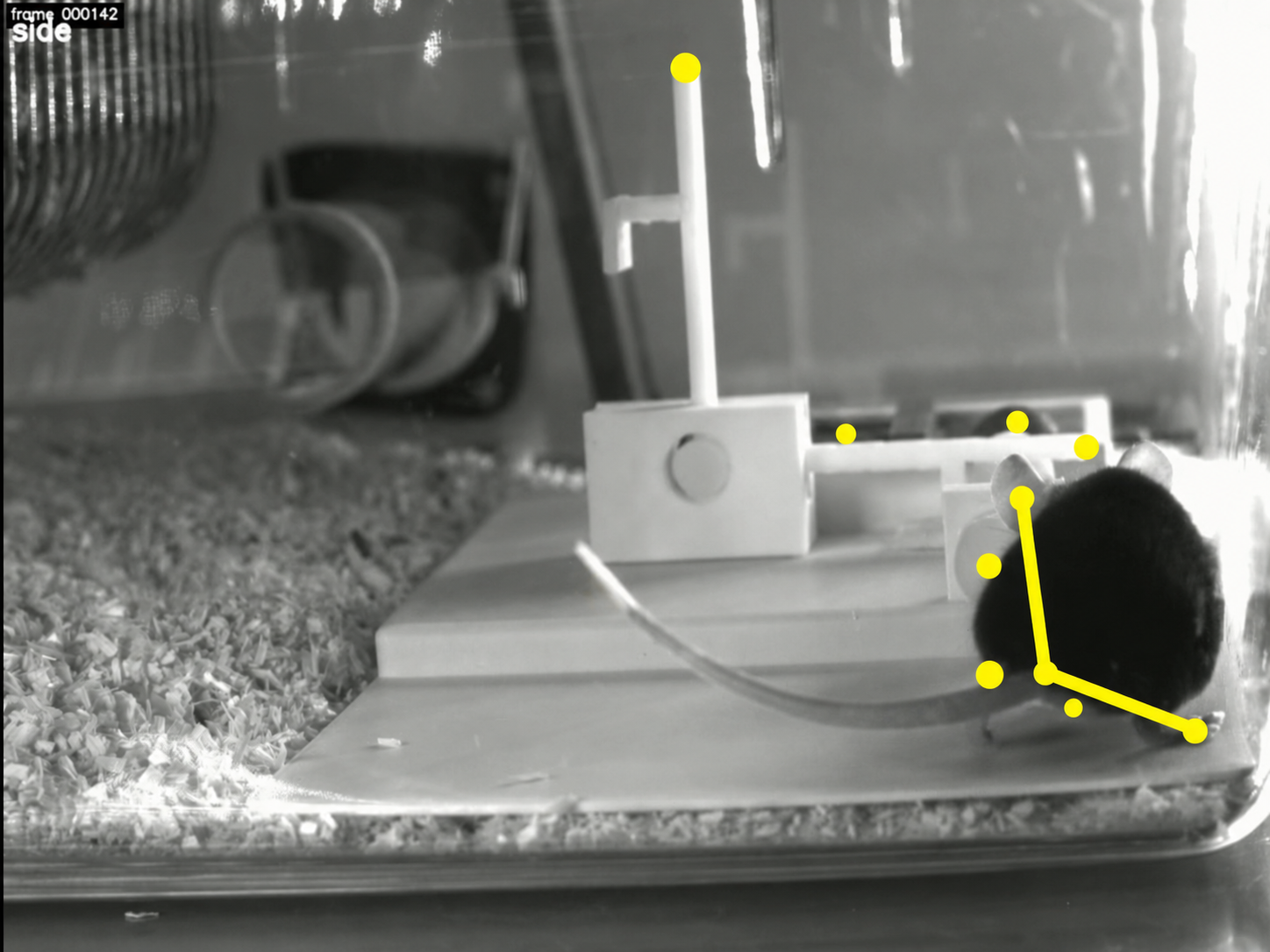}
  \end{minipage}\hspace{0.05\textwidth}
  \begin{minipage}[t]{0.35\textwidth}
    \centering
    \includegraphics[width=\textwidth]{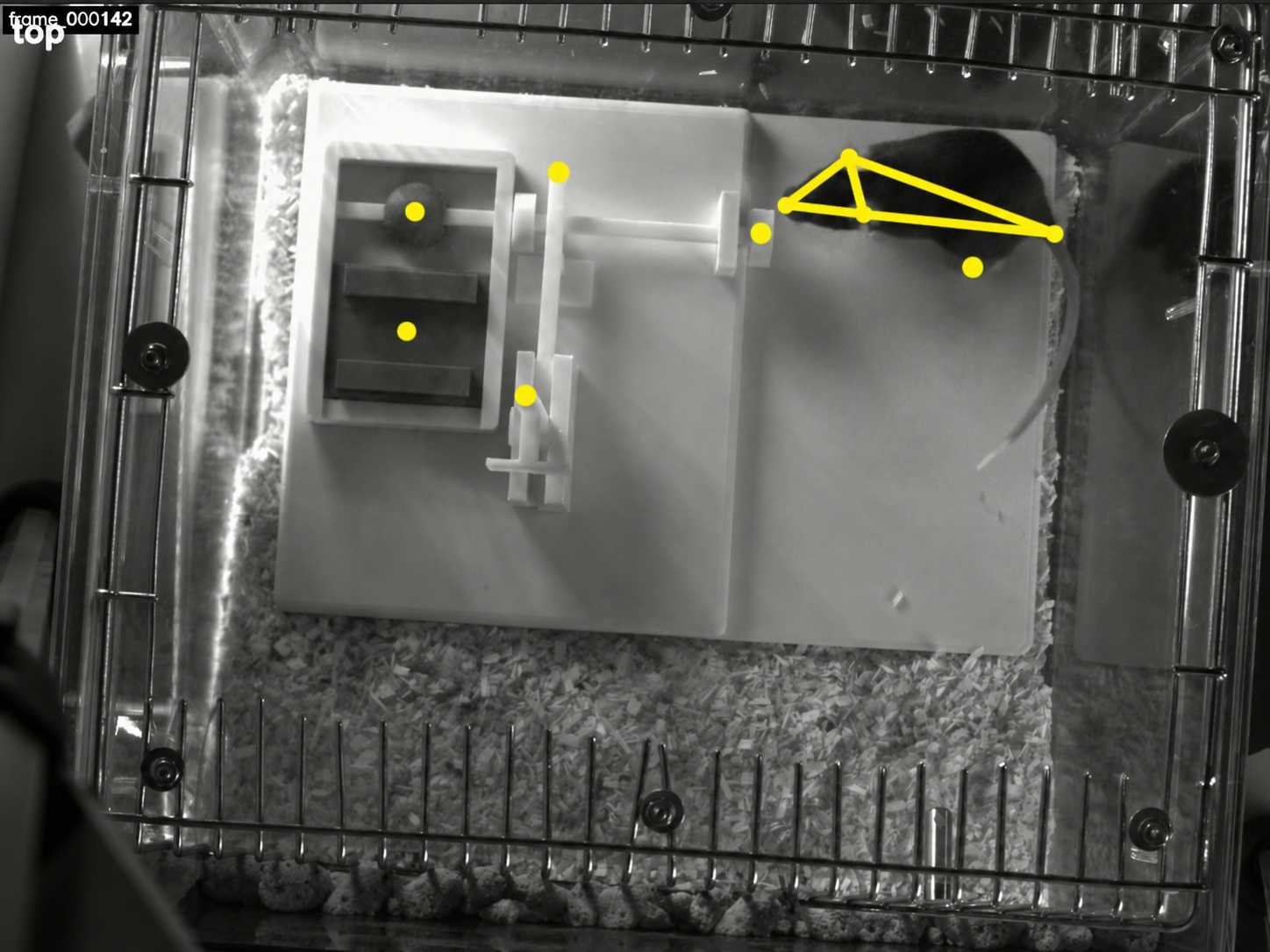}\\[0.5em]
    \includegraphics[width=\textwidth]{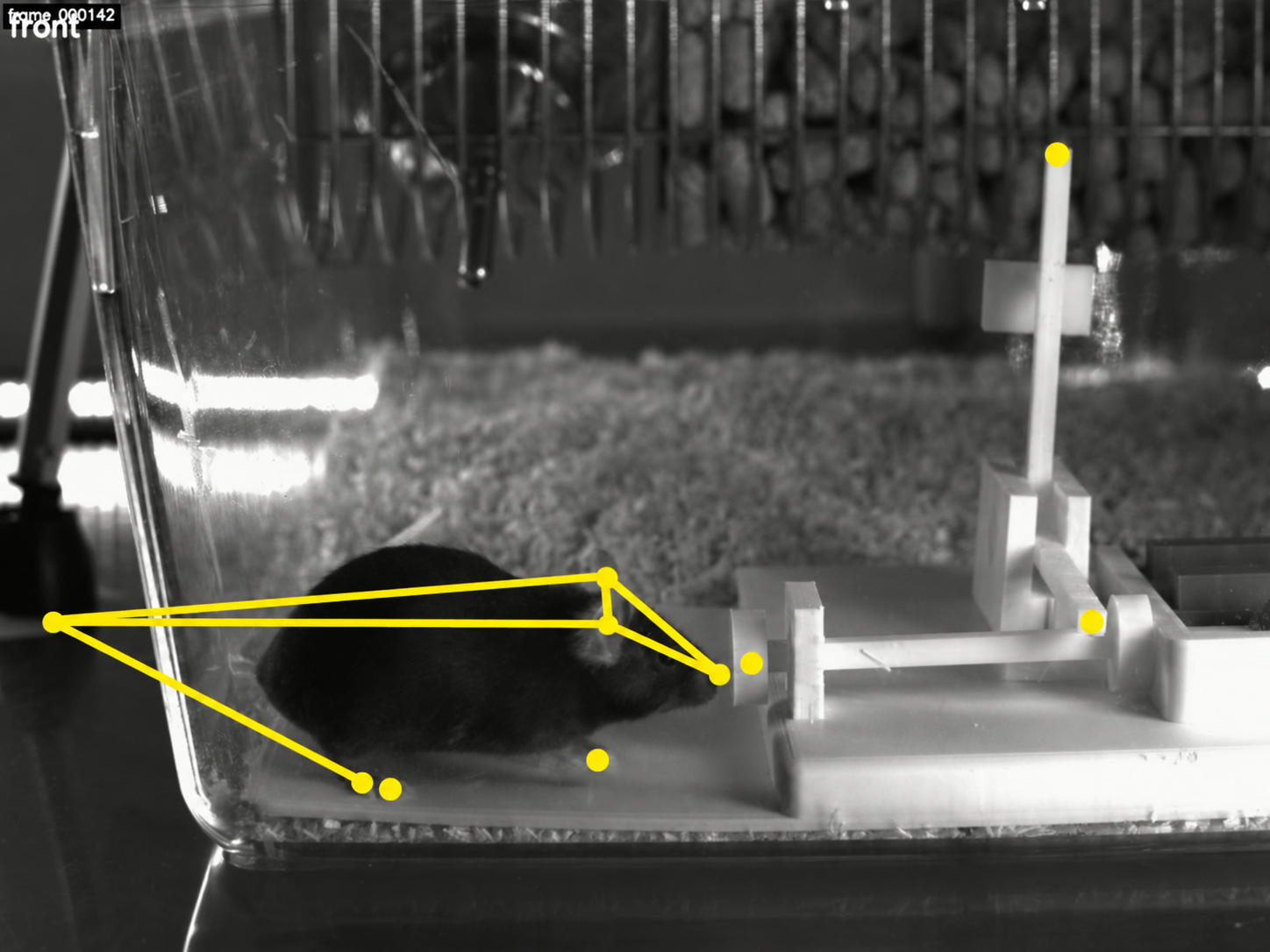}
  \end{minipage}
  \vspace{6pt}
  \caption{Mouse lock box experiment. The tail root in the front view is wrongly extracted.}
  \label{fig:Lockbox}
\end{figure*}

The lock-box task requires rotating a lever, removing a splint, moving a ball, and shifting a lid to reach an oat flake reward. The data set of \cite{boon2024mechanical} provides DeepLabCut-extracted body and lock-box keypoints \cite{mathis2018deeplabcut}; despite methodological differences from Example A, similar cleaning is required (Fig. \ref{fig:Lockbox}).

\subsection{Calibrating the Cameras With the Lock-Box}

\begin{figure*}[!tb]
  \centering
  \includegraphics[height=3.5cm]{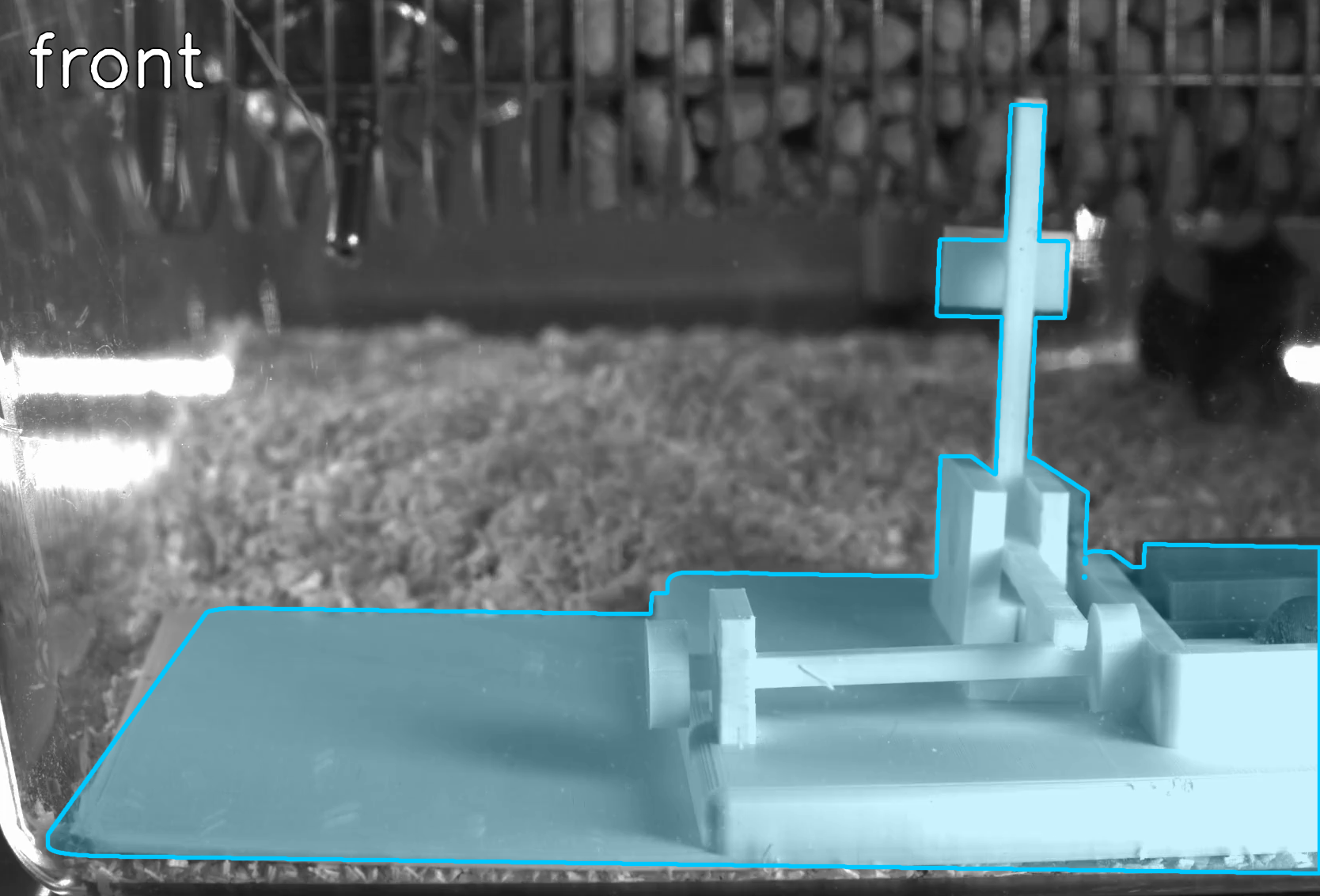}\hfill
  \includegraphics[height=3.5cm]{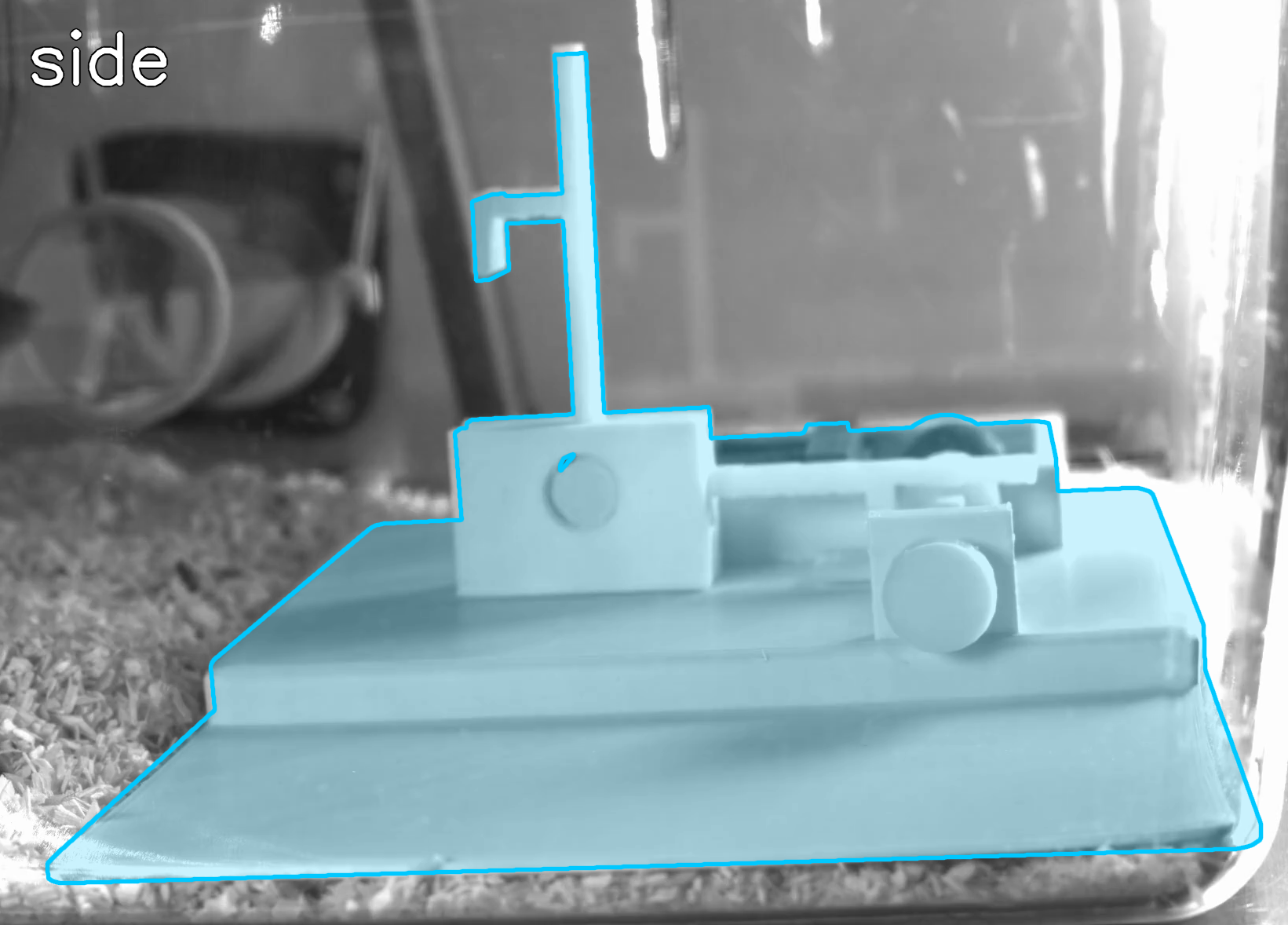}\hfill
  \includegraphics[height=3.5cm]{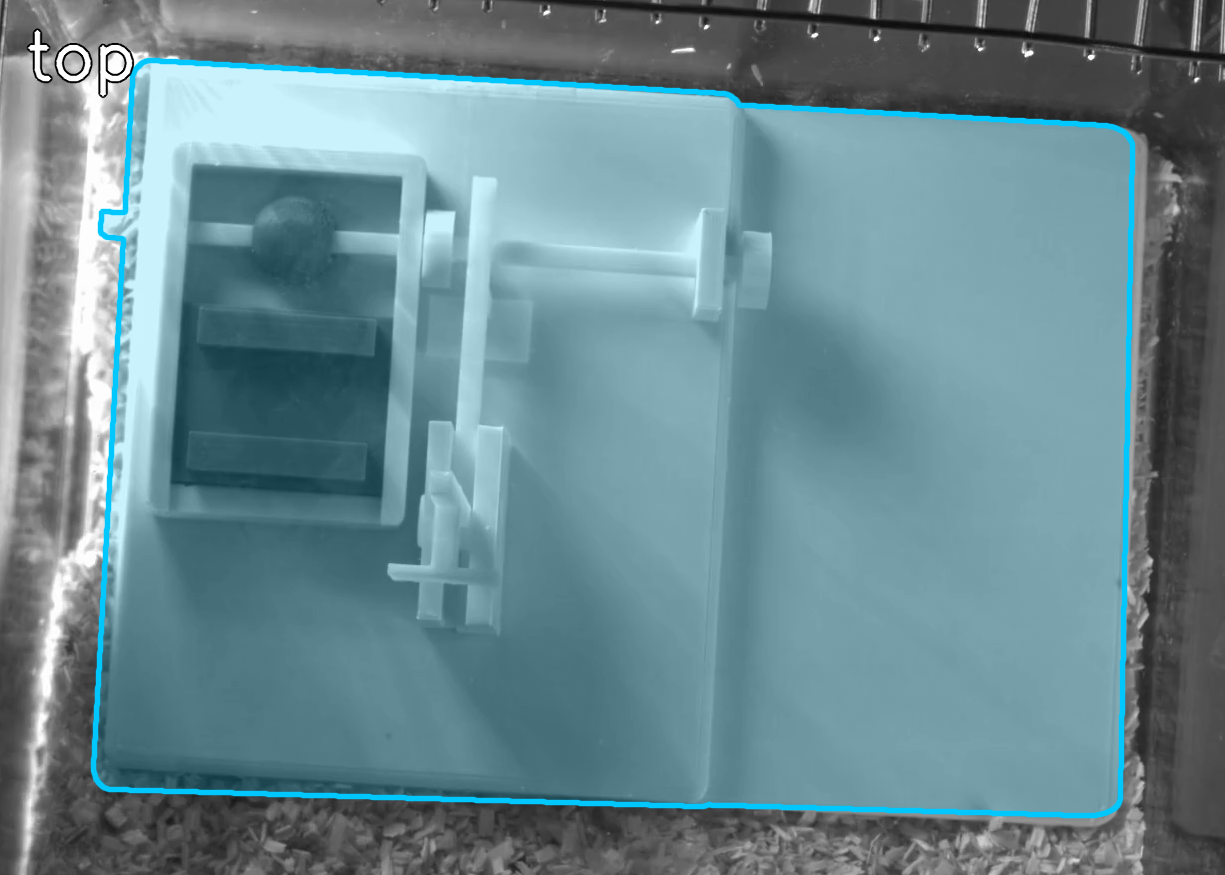}
  \vspace{6pt}
  \caption{Camera calibration with the lock-box. The lock-box CAD model ({\color[RGB]{0,200,255}blue}), projected with the fitted camera parameters, covers the lock-box in the front, side, and top views.}
  \label{fig:LBcalibration}
\end{figure*}

Combining the three views requires the position, orientation, and focal length of each camera. These are usually determined from images of a calibration pattern. Here, the lock-box takes this role, since its exact 3D shape is known from its CAD model. For each camera, the parameters are chosen such that the projected CAD model covers the lock-box in the image (Fig. \ref{fig:LBcalibration}).

Fitting each image separately is not sufficient: a single view leaves trade-offs, e.g. between focal length and camera distance, and the three separately fitted cameras disagree on where lock-box landmarks such as the lever tip lie in 3D. The cameras are therefore refined jointly, additionally requiring that the lock-box keypoints detected by DeepLabCut be consistent across views. The measured width of the lock-box base (199\,mm) sets the metric scale. The lock-box also defines the coordinate frame, so mouse positions are expressed directly relative to the puzzle.

\subsection{From 2D Keypoints to 3D Keypoints}

A keypoint detected in one image defines a viewing ray in space; the rays of two cameras intersect at the keypoint's 3D position (triangulation). With three cameras, any two views suffice to compute a 3D point, and the third view provides an independent check: the 3D point is projected into it and compared with the detection there. A point is accepted if all three views agree. If only two views provide reliable detections, the point is accepted but marked as unchecked. If the third view contradicts the other two, the point is rejected, as are points implying implausible bone lengths or jumps between frames. Rejected points are not replaced by guesses; only gaps of up to five frames (0.17\,s) are interpolated and marked as such.

\begin{figure*}[!tb]
  \centering
  \includegraphics[height=4.6cm]{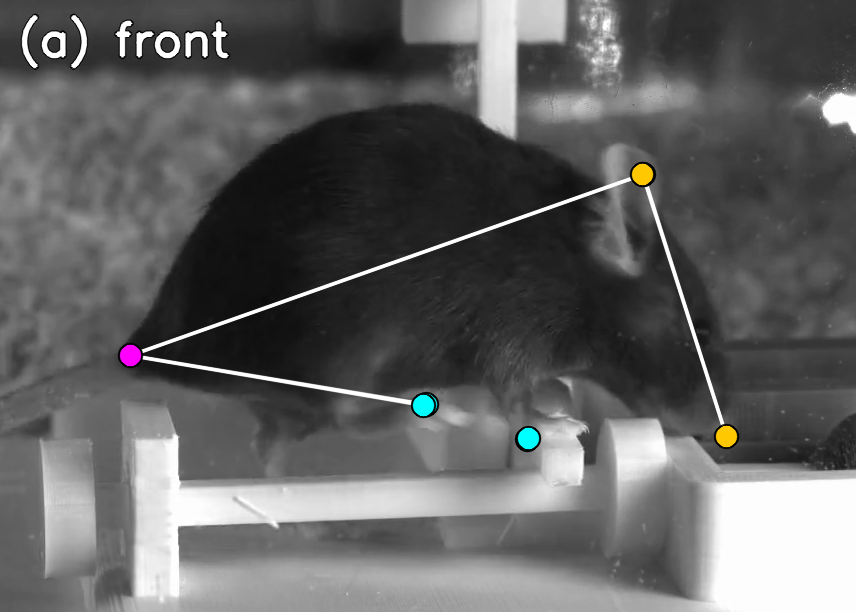}\hspace{0.02\textwidth}%
  \includegraphics[height=4.6cm]{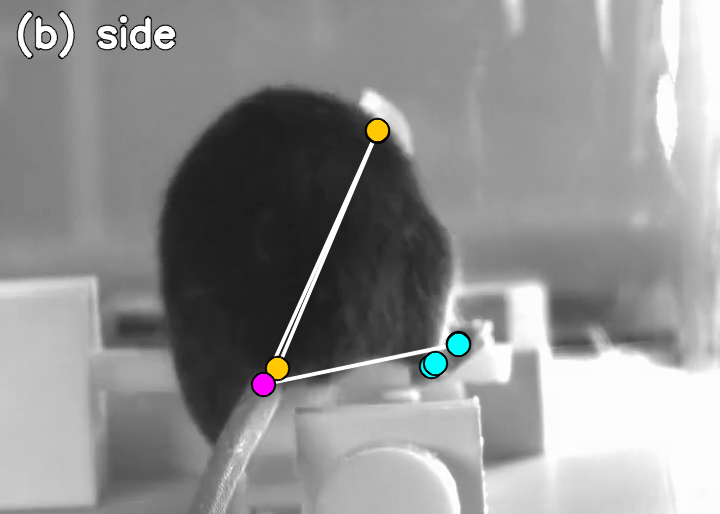}\\[0.5em]
  \includegraphics[height=4.6cm]{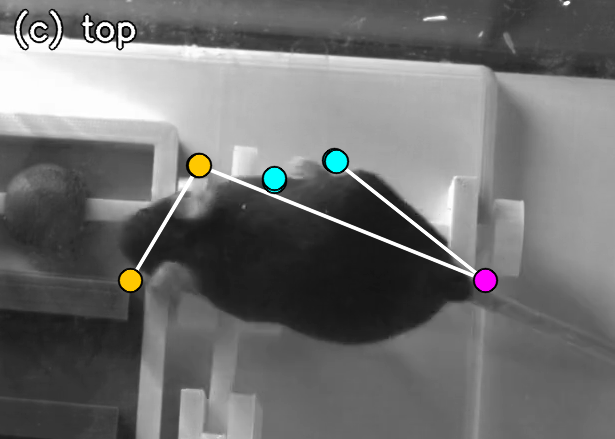}\hspace{0.02\textwidth}%
  \includegraphics[height=4.6cm]{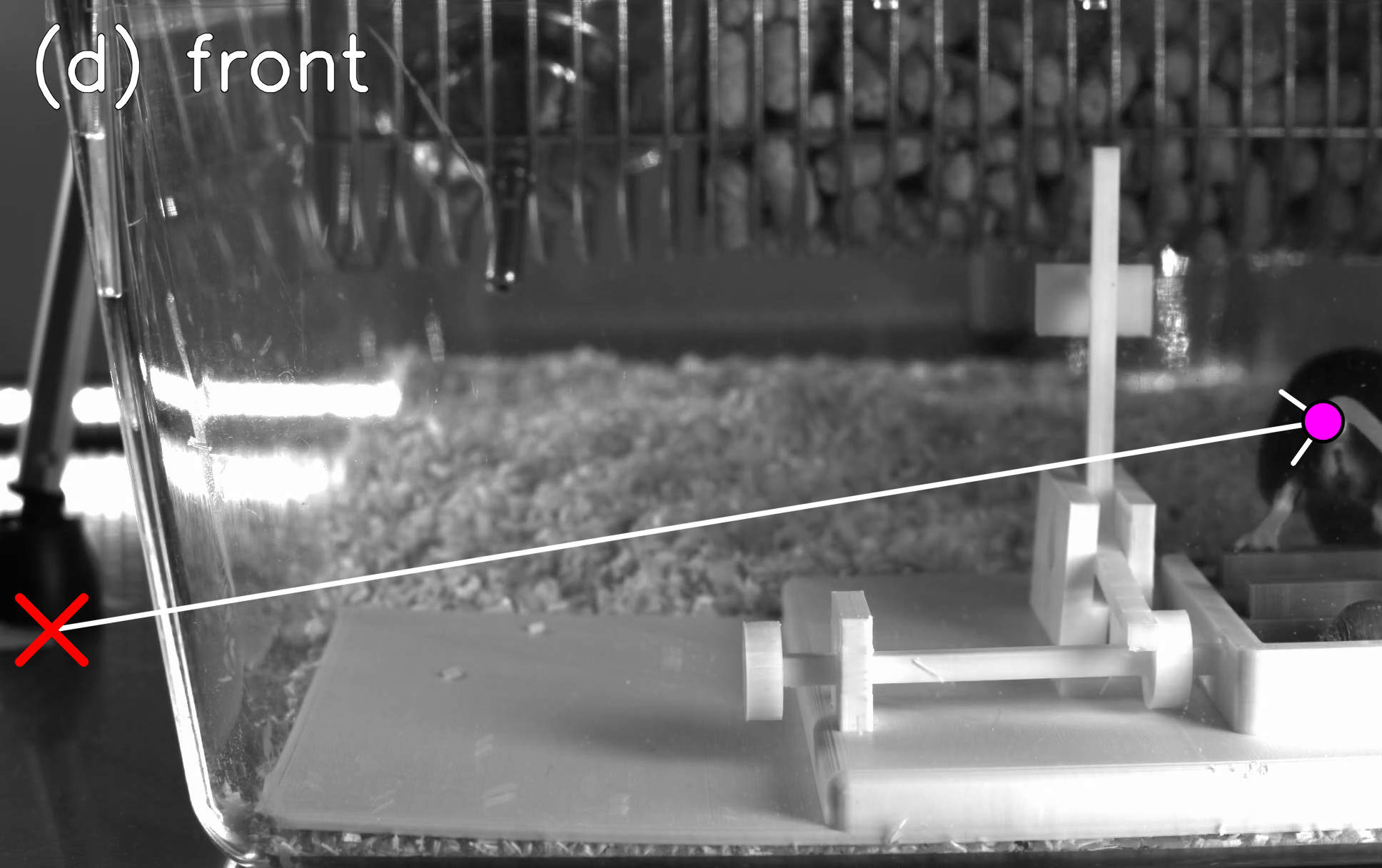}
  \vspace{6pt}
  \caption{3D reconstruction of the mouse. (a)--(c) Keypoints reconstructed in 3D from all three views and projected back into each image: head (nose, ears) {\color[RGB]{255,200,0}$\bullet$}, paws {\color[RGB]{0,220,220}$\bullet$}, tail root {\color[RGB]{255,0,255}$\bullet$}; white lines connect neighboring keypoints of the skeleton. (d) A keypoint error revealed by geometry: DeepLabCut places the tail root at the red cross (likelihood 0.95) at the left image border. Side and top views agree on a tail-root position whose projection into the front view ({\color[RGB]{255,0,255}$\bullet$}) lies on the mouse, about 1,800 pixels away.}
  \label{fig:LBmouse}
\end{figure*}

Fig. \ref{fig:LBmouse}(a)--(c) shows 3D keypoints reconstructed in this way and projected back into the three images: in every view they lie on the corresponding body parts.

\subsection{Keypoint Errors Revealed by the Third View}

Across the recording, roughly a third of all mouse keypoint detections are inconsistent with the other views, and many of these are confident detector errors. In about 40\,\% of the frames in which DeepLabCut reports the tail root with a likelihood above 0.9 in all three views, the three detections cannot belong to a single 3D point. Typically, the side and top views agree while the front-view detection lies far from the mouse (Fig. \ref{fig:Lockbox} and Fig. \ref{fig:LBmouse}(d)). Within the front view alone, nothing flags such an error, as the detector reports high confidence. A classifier operating directly on 2D keypoints therefore receives these errors as valid input, whereas the 3D reconstruction detects and removes them through the geometric constraint between views.

\subsection{Accuracy of the Reconstruction}

The real recording has no 3D ground truth, so accuracy cannot be measured directly. We therefore estimated it in a simulation built from the real data: the reconstructed mouse motion is treated as known ground truth, projected into the three calibrated cameras, perturbed with the detection noise and missing detections observed in the real recording, and reconstructed again with the same pipeline. The median 3D error is about 2\,mm for head and tail root and about 4\,mm for the paws. If the cameras are additionally misaligned by $2^\circ$, the errors rise to about 14\,mm, so the reconstruction is most sensitive to camera calibration. These results hold for one recording and the assumed noise model.

\section{Conclusion}

The examples support three conclusions:
\begin{itemize}
    \item Preferably, behavior should not be directly estimated from video observation, but first 3D reconstruction should be conducted, and then behavior should be inferred from 3D estimates. This does not exclude replacing explicit 3D reconstruction with implicit neural representation.
    \item Behavior analysis of real animals profits from parallel full simulation of the animals' behavior, video acquisition, 3D reconstruction, and behavior estimation.
    \item Incorporating static and dynamic 3D model-based constraints, along with motion constraints, into the traditional image observation constraints enhances 3D reconstruction from video data.
\end{itemize}

Behavior classification approaches using a single video camera to observe the animals can be integrated into our concept by "depth from single image" networks and geometric 3D models providing constraints for 3D reconstruction.

Following our strategy does not exclude later switching to end-to-end deep-learning strategies that incorporate e.g. implicit 3D representations.

\subsubsection*{Acknowledgments}
Funded by the Deutsche Forschungsgemeinschaft (DFG, German Research Foundation) under Germany’s Excellence Strategy – EXC 2002/1 “Science of Intelligence” – project number 390523135, and through the Research Training Group GRK 3051-,,3RTG”.
%
%
%
\bibliographystyle{ieeetr}
\bibliography{references}

@inproceedings{hellwich2000object,
  title={Object extraction from high-resolution multisensor image data},
  author={Hellwich, Olaf and Wiedemann, Christian},
  booktitle={Third International Conference Fusion of Earth Data, Sophia Antipolis},
  volume={115},
  year={2000}
}

@article{lauer2022multi,
  title={Multi-animal pose estimation, identification and tracking with {DeepLabCut}},
  author={Lauer, Jessy and Zhou, Mu and Ye, Shaokai and Menegas, William and Schneider, Steffen and Nath, Tanmay and Rahman, Mohammed Mostafizur and Di Santo, Valentina and Soberanes, Daniel and Feng, Guoping and others},
  journal={Nature Methods},
  volume={19},
  number={4},
  pages={496--504},
  year={2022},
  publisher={Nature Publishing Group US New York}
}

@inproceedings{ronneberger2015unet,
  title     = {{U-Net}: Convolutional Networks for Biomedical Image Segmentation},
  author    = {Ronneberger, Olaf and Fischer, Philipp and Brox, Thomas},
  booktitle = {Medical Image Computing and Computer-Assisted Intervention -- MICCAI 2015},
  pages     = {234--241},
  year      = {2015},
  publisher = {Springer},
  doi       = {10.1007/978-3-319-24574-4_28}
}

@article{mars2021,
  author = {Segalin, Cristiano and Williams, Jonathan and Karigo, Tomomi and Hui, Min and Zelikowsky, Michelle and Sun, Jenny J. and Perona, Pietro and Anderson, David J.},
  title = {The Mouse Action Recognition System ({MARS}) software pipeline for automated analysis of social behaviors in mice},
  journal = {eLife},
  volume = {10},
  pages = {e63720},
  year = {2021},
  doi = {10.7554/eLife.63720}
}

@article{bsoid2021,
  author = {Hsu, Alexander I. and Yttri, Eric A.},
  title = {{B-SOiD}, an open-source unsupervised algorithm for identification and fast prediction of behaviors},
  journal = {Nature Communications},
  volume = {12},
  number = {5188},
  year = {2021},
  doi = {10.1038/s41467-021-25420-x}
}

@article{vame2022,
  author = {Luxem, Konrad and Fuhrmann, Fabian and Kürsch, Jonas and Remy, Simon and Bauer, Petra and Schubert, David and Kühn, Alexander A.},
  title = {Identifying behavioral structure from deep variational embeddings of animal motion},
  journal = {Communications Biology},
  volume = {5},
  number = {600},
  year = {2022},
  doi = {10.1038/s42003-022-04080-7}
}

@article{simba2024,
  author = {Nilsson, Simon R.O. and Goodwin, Nina L. and Choong, Jie J. and Hwang, Sophia and Wright, Hannah R. and Norville, Zeb T. and Serreau, Pauline and Peron, Simon and Golden, Sam A.},
  title = {Simple Behavioral Analysis ({SimBA}) -- an open source toolkit for computer classification of complex social behaviors in experimental animals},
  journal = {Nature Neuroscience},
  volume = {27},
  pages = {1003--1015},
  year = {2024},
  doi = {10.1038/s41593-024-01649-9}
}

@article{keypointmoseq2024,
  author = {Weinreb, Caleb and Sosa-Plata, Cristian and Bendesky, Andres and Datta, Sandeep R.},
  title = {Keypoint-{MoSeq}: parsing behavior by linking point tracking to pose dynamics},
  journal = {Nature Methods},
  volume = {21},
  pages = {1146--1156},
  year = {2024},
  doi = {10.1038/s41592-024-02318-2}
}

@article{asoid2024,
  author = {Hsu, Alexander I. and Yttri, Eric A.},
  title = {{A-SOiD}: active learning for expert-guided discovery of behavior},
  journal = {Nature Methods},
  volume = {21},
  pages = {597--605},
  year = {2024},
  doi = {10.1038/s41592-024-02200-1}
}

@article{dlc2action2025,
  author = {Kozlova, Elizaveta and Bonnetto, Andy and Matthis, Alexander},
  title = {{DLC2Action}: Deep Learning for Animal Action Recognition from Pose Trajectories},
  journal = {bioRxiv},
  year = {2025},
  note = {Preprint},
  doi = {10.1101/2025.09.27.678941}
}

@article{mathis2018deeplabcut,
  title={{DeepLabCut}: markerless pose estimation of user-defined body parts with deep learning},
  author={Mathis, Alexander and Mamidanna, Pranav and Cury, Kevin M and Abe, Taiga and Murthy, Venkatesh N and Mathis, Mackenzie Weygandt and Bethge, Matthias},
  journal={Nature neuroscience},
  volume={21},
  number={9},
  pages={1281--1289},
  year={2018},
  publisher={Nature Publishing Group US New York}
}

@article{boon2024mechanical,
  title={Mechanical problem solving in mice},
  author={Boon, Marcus N and Andresen, Niek and Traverso, Soledad and Meier, Sophia and Schuessler, Friedrich and Hellwich, Olaf and Lewejohann, Lars and Th{\"o}ne-Reineke, Christa and Sprekeler, Henning and Hohlbaum, Katharina},
  journal={bioRxiv},
  pages={2024--07},
  year={2024},
  publisher={Cold Spring Harbor Laboratory}
}

@misc{Agarwal_Ceres_Solver_2022,
  author = {Agarwal, Sameer and Mierle, Keir and The Ceres Solver Team},
  title = {{Ceres Solver}},
  license = {Apache-2.0},
  url = {https://github.com/ceres-solver/ceres-solver},
  version = {2.2},
  year = {2023},
  month = {10}
}

@article{hellwich2024tracking,
  title={Tracking Mouse from Incomplete Body-Part Observations and Deep-Learned Deformable-Mouse Model Motion-Track Constraint for Behavior Analysis},
  author={Hellwich, Olaf and Andresen, Niek and Hohlbaum, Katharina and Boon, Marcus N and Kwiatkowski, Monika and Matern, Simon and Reiske, Patrik and Sprekeler, Henning and Th{\"o}ne-Reineke, Christa and Lewejohann, Lars and Ghani Zada, Huma and Brück, Michael and Traverso, Sole},
  journal={Festschrift f{\"u}r Prof. Dr.-Ing. Helmut Mayer zum 60. Geburtstag},
  pages={45--52},
  year={2024}
}
\end{document}